\documentclass[10pt,twocolumn,letterpaper]{article}

\usepackage{wacv}              

\usepackage{float} 
\usepackage{graphicx}

\definecolor{wacvblue}{rgb}{0.21,0.49,0.74}
\usepackage[pagebackref,breaklinks,colorlinks,allcolors=wacvblue]{hyperref}

\def\wacvPaperID{2} 
\def\confName{WACV}
\def\confYear{2026}

\title{Few-Shot LoRA Adaptation of a Flow-Matching Foundation Model for Cross-Spectral Object Detection}

\author{
Maxim Clouser \qquad Kia Khezeli \qquad John Kalantari\\[0.75em]
Yrikka Inc.\\
{\tt\small max@yrikka.com, kia@yrikka.com, john@yrikka.com}
}

\begin{document}
\maketitle
\begin{abstract}
Foundation models for vision are predominantly trained on RGB data, while many safety-critical applications rely on non-visible modalities such as infrared (IR) and synthetic aperture radar (SAR). We study whether a single flow-matching foundation model pre-trained primarily on RGB images can be repurposed as a cross-spectral translator using only a few co-measured examples, and whether the resulting synthetic data can enhance downstream detection. Starting from FLUX.1 Kontext, we insert low-rank adaptation (LoRA) modules and fine-tune them on just 100 paired images per domain for two settings: RGB→IR on the KAIST dataset and RGB→SAR on the M4-SAR dataset. The adapted model translates RGB images into pixel-aligned IR/SAR, enabling us to reuse existing bounding boxes and train object detection models purely in the target modality. Across a grid of LoRA hyperparameters, we find that LPIPS computed on only 50 held-out pairs is a strong proxy for downstream performance: lower LPIPS consistently predicts higher mAP for YOLOv11n on both IR and SAR, and for DETR on KAIST IR test data. Using the best LPIPS-selected LoRA adapter, synthetic IR from external RGB datasets (LLVIP, FLIR ADAS) improves KAIST IR pedestrian detection, and synthetic SAR significantly boosts infrastructure detection on M4-SAR when combined with limited real SAR. Our results suggest that few-shot LoRA adaptation of flow-matching foundation models is a promising path toward foundation-style support for non-visible modalities.
\end{abstract}
    
\section{Introduction}

The rapid progress of foundation models in computer vision has been largely confined to the
visible spectrum, where massive datasets of RGB images underpin training of general-purpose
models like CLIP and Stable Diffusion~\cite{radford2021clip,rombach2022ldm}. By contrast, many real-world applications in infrared
(IR) and synthetic aperture radar (SAR) domains lack comparable data scale and pre-trained models~\cite{chowdhury2023fomo}. This gap
is critical: autonomous driving, surveillance, and remote sensing all require perception beyond
visible light, yet current vision foundation models struggle to generalize to modalities such as IR or SAR. The motivation for this work arises from the need to extend foundation models beyond the visible spectrum, leveraging their powerful learned representations to benefit
low-resource sensing domains.

One promising strategy is cross-spectral image translation, using models trained on abundant visible imagery to synthesize corresponding IR or SAR views. Prior GAN-based translators can produce realistic outputs but require dataset-specific training and can be unstable across spectra, sometimes hallucinating structures or distorting IR brightness~\cite{isola2017pix2pix,zhu2017cyclegan,song2019thermalgap,qiu2020thermalcycle,zhang2021improvedcycle}. Diffusion-based translation can be more stable~\cite{ho2020ddpm,huang2023vidiff}, but training a diffusion model from scratch for each sensor pair remains costly~\cite{li2023diffusionsurvey}. We therefore ask whether a single pre-trained foundation generator can be repurposed into a reusable cross-spectral translator with only a small number of co-measured examples. Concretely, we adapt FLUX.1 Kontext~\cite{lipman2023flowmatching,flux2025kontext}, a latent rectified-flow transformer pre-trained on large-scale RGB data, by inserting Low-Rank Adaptation (LoRA) modules~\cite{hu2022lora} and fine-tuning only these parameters on a small paired set of aligned RGB--IR or RGB--SAR examples. This yields a parameter-efficient mapping that preserves the base model prior while keeping data and compute requirements low.

We validate our approach on two datasets: KAIST multispectral (RGB--IR pedestrian scenes)~\cite{hwang2015kaist} and M4-SAR (an RGB--SAR satellite imagery benchmark for object detection)~\cite{chao2023m4sar}. With as few as 100 paired examples per domain, LoRA-adapted FLUX.1 Kontext generates realistic IR or SAR images that are useful for downstream detection. We find that LPIPS (Learned Perceptual Image Patch Similarity) scores between synthetic and real images~\cite{zhang2018lpips} strongly correlates with target-domain mAP: lower LPIPS on a small validation set reliably predicts better detector performance. Using the best LPIPS-selected LoRA adapter, we show two practical uses of cross-spectral augmentation: (1) translating RGB-only LLVIP and FLIR ADAS images into the KAIST IR domain, which improves KAIST pedestrian detection over training on limited real IR alone; and (2) augmenting M4-SAR with synthetic SAR generated from co-registered RGB images, yielding a notable mAP gain over using only real SAR. In summary, our contributions include the following:

\begin{itemize}
\item \textbf{Parameter-efficient cross-spectral translation.} We adapt a single flow-matching diffusion foundation model for cross-spectral image translation (RGB$\rightarrow$IR, RGB$\rightarrow$SAR) via LoRA fine-tuning. With only 100 paired examples per dataset, the resulting adapters produce high-fidelity translations and act as reusable cross-spectral translators, illustrating an effective way to extend vision foundation models beyond the visible spectrum with modest data and compute.
\item \textbf{Correlation of perceptual quality with detection.} We show that LPIPS perceptual similarity~\cite{zhang2018lpips} on a small validation set is a reliable indicator of downstream utility: lower LPIPS correlates with higher detection mAP on both IR and SAR tasks, supporting its use as a proxy metric when labeled detection data are scarce.
\item \textbf{Boosting detection with cross-spectral data augmentation.} We translate additional RGB datasets into the target modality and reuse their labels to augment detector training, improving performance in low-data IR and SAR settings.
\end{itemize}

\section{Related Work}

\subsection{Cross-Spectral Image Translation}

Early approaches to cross-spectral translation employed generative adversarial networks.
Pix2pix demonstrated supervised translation using paired images and a conditional GAN
objective~\cite{isola2017pix2pix}. To relax the need for pairing, CycleGAN introduced cycle-consistency losses
enabling translation between unpaired datasets~\cite{zhu2017cyclegan}. These frameworks inspired numerous
extensions: UNIT and MUNIT incorporated stochastic mappings for multimodal outputs~\cite{liu2017unit,huang2018munit},
and many domain-specific GAN models have been proposed for spectral translation. For
RGB$\rightarrow$IR translation, specialized GANs such as ThermalGAN~\cite{kniaz2018thermalgan} are used to generate thermal images for person re-identification, and InfraGAN~\cite{ren2022infragan} is used to improve realism of IR outputs. IR translation has also been studied with attention to stability and detail. For example,
Ma et al. fuse multi-scale features in a pix2pix-based IR generator, and other works introduce
architectural variants to better translate thermal face images to visible, all within GAN-style
frameworks~\cite{kniaz2018thermalgan,ren2022infragan,song2019thermalgap,qiu2020thermalcycle,zhang2021improvedcycle}.

In remote sensing, RGB--SAR translation techniques based on CycleGAN are used to compensate for sensing gaps, e.g., generating RGB-like images from SAR when clouds
obscure satellite imagery~\cite{wang2020sar2opt}. Variants of CycleGAN and related architectures introduce
additional structural constraints such as segmentation-guided losses to maintain object shapes,
road topology, or coastline structure during SAR$\rightarrow$RGB translation~\cite{wang2020sar2opt,song2019thermalgap,qiu2020thermalcycle,zhang2021improvedcycle}. Despite these
advances, GAN-based cross-spectral translation remains challenging. Typical failure modes
include misalignment of fine details, brightness distortions, and hallucinated structures in the
generated images~\cite{song2019thermalgap,qiu2020thermalcycle,zhang2021improvedcycle}.

Modern diffusion models offer a compelling alternative due to their stability and sample quality~\cite{ho2020ddpm}. Denoising diffusion and score-based models have been applied to cross-spectral tasks.
For instance, VI-Diff employs a diffusion model for unpaired RGB$\rightarrow$IR translation in person
re-identification~\cite{huang2023vidiff}. VI-Diff reports improved fidelity of synthetic IR for re-identification, albeit with high computational cost and dataset-specific training~\cite{huang2023vidiff,li2023diffusionsurvey}. Physics-Informed Diffusion (PID) builds on HADAR-style physics-based IR formation~\cite{bao2023heat} and introduces a TeV decomposition together with physics-informed reconstruction and TeV-space consistency losses for physically grounded IR image translation~\cite{mao2026pid}. Our work differs in that we do not train a diffusion model from scratch for each spectral pair. Instead, we fine-tune a general pre-trained model. By leveraging a strong diffusion model trained on large-scale RGB data
\cite{rombach2022ldm,lipman2023flowmatching,flux2025kontext}, we obtain excellent cross-spectral translation with only a fraction of the data and training time that a bespoke GAN or diffusion would require.

\subsection{LoRA and Parameter-Efficient Fine-Tuning of Generative Models}

Low-Rank Adaptation (LoRA)~\cite{hu2022lora} injects small trainable low-rank matrices into a pre-trained network, enabling parameter-efficient fine-tuning without modifying most of the original weights. Originally proposed for adapting large language models, LoRA has since been widely used to customize diffusion-based image generators such as Stable Diffusion to new styles or concepts with modest compute~\cite{rombach2022ldm,hu2022lora,li2023diffusionsurvey}. We follow this paradigm and adapt the FLUX.1 Kontext model to IR and SAR translation using LoRA adapters that comprise less than 1\% of the base model parameters, yet suffice to imprint the cross-spectral mapping in data-scarce regimes.

\subsection{Flow-Matching and Score-Based Generative Models in Vision}

Score-based diffusion models such as DDPMs generate images via iterative denoising and have revolutionized image synthesis~\cite{ho2020ddpm}. Flow matching generalizes this family by training continuous normalizing flows to match a time-indexed probability flow between noise and data, reproducing diffusion behavior while improving stability~\cite{lipman2023flowmatching}. Rectified flow transformers further scale these ideas to high-resolution image generation with competitive quality and efficient sampling~\cite{chen2023rectifiedflow}. We leverage FLUX.1 Kontext, a latent rectified-flow model that supports flexible conditioning and in-context image editing by accepting both an input image and a text prompt~\cite{flux2025kontext,esser2023dct}. Its unified training on generation and editing tasks~\cite{flux2025kontext,esser2023dct} makes FLUX.1 a natural backbone for cross-spectral translation.

\subsection{Synthetic IR/SAR Data for Object Detection}

Prior work has used synthetic images to improve IR and SAR object detection when real annotations are scarce. For infrared pedestrian detection, several works use CycleGAN-style RGB$\rightarrow$IR translation while reusing RGB bounding boxes, improving IR detectors without extra thermal labels~\cite{song2019thermalgap,qiu2020thermalcycle,zhang2021improvedcycle}. In SAR, augmentation by translation is less explored, but M4-SAR shows that combining RGB and SAR inputs improves detection over SAR alone, suggesting that RGB imagery provides complementary structure~\cite{chao2023m4sar}. Our work follows this line by generating synthetic IR and SAR with a foundation model and using them as additional training data. Unlike prior GAN-based approaches, we leverage a single LoRA-adapted flow-matching model rather than training task-specific translators from scratch.

\section{Method: LoRA-Adapted Flow Matching for Cross-Spectral Translation}

\subsection{Problem Setting}

We consider cross-spectral translation between a source modality \(x^{s}\) (RGB) and a target modality \(x^{t}\) (IR or SAR). Given a small set of aligned pairs
\[
\mathcal{D}_\text{pair} = \{(x^{s}_i, x^{t}_i)\}_{i=1}^{N},
\]
and a larger set of labeled target-domain images \(\mathcal{D}_\text{det}\) with bounding boxes \(\mathcal{B}\), our goal is to:

\begin{enumerate}
    \item Learn a conditional generator \(G_\theta\) that translates \(x^{s}\) into a synthetic target image \(\hat{x}^{t} = G_\theta(x^{s})\) that is pixel-aligned with \(x^{s}\).
    \item Use synthetic images \(\hat{x}^{t}\) to train object detectors that operate purely in the target domain (IR or SAR), either by augmenting \(\mathcal{D}_\text{det}\) with synthetic data or by training on synthetic target images alone.
\end{enumerate}

\begin{figure*}[t]
    \centering
    \includegraphics[width=0.8\textwidth]{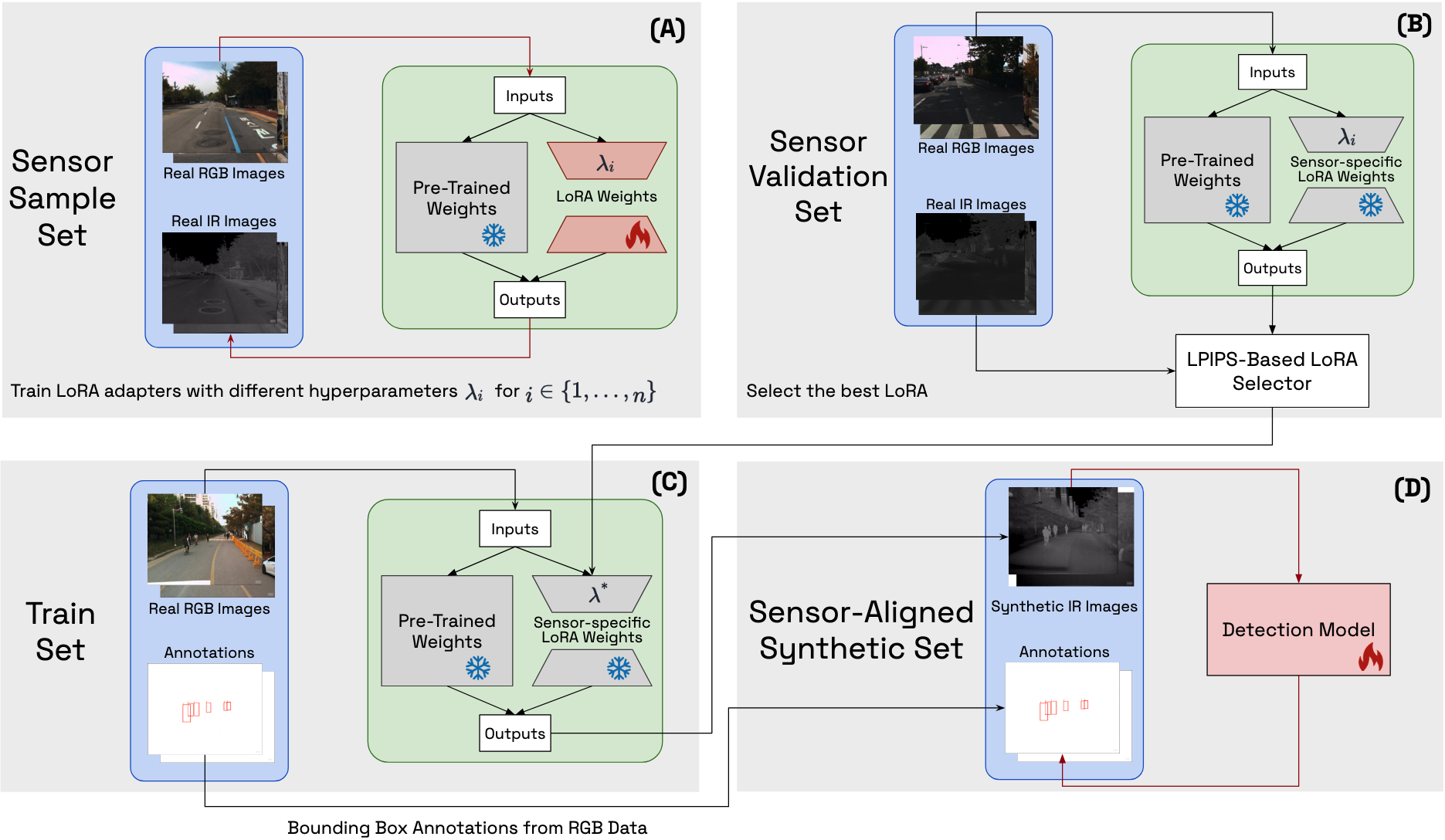}
    \caption{
    Overview of our pipeline for LoRA-adapted flow-matching cross-spectral translation and detection.
    (A) A small \emph{Sensor Sample} split of paired RGB and IR/SAR images is used to train multiple
    LoRA configurations on top of a frozen FLUX.1 Kontext base model.
    (B) A separate \emph{Sensor Val} split is translated and scored against real images with LPIPS to select the best LoRA.
    (C) The selected LoRA \(\lambda^\ast\) is applied to the RGB \emph{Train} split to generate a sensor-aligned synthetic target-domain set, reusing the original RGB bounding box annotations.
    (D) The sensor-aligned synthetic set, optionally combined with real target-domain images, is used to train an object detection model in the target modality.
    All panels show KAIST RGB$\rightarrow$IR; the same pipeline is applied for RGB$\rightarrow$SAR.
    }
    \label{fig:pipeline}
\end{figure*}

The main constraint is that \(|\mathcal{D}_\text{pair}|\) is very small (100 pairs per dataset in our experiments) and is used exclusively to train the LoRA adapters, reflecting the scarcity of co-measured cross-spectral data. Figure~\ref{fig:pipeline} summarizes the overall pipeline from LoRA adaptation to detector training. We denote the small paired subset used for LoRA training as the \emph{Sensor Sample} split, a disjoint paired subset for validation as \emph{Sensor Val}, and the full detection training images as the \emph{Train} split.

\subsection{Base Model: FLUX.1 Kontext}

We build on FLUX.1 Kontext, a 12B-parameter rectified-flow transformer trained in the latent space of an autoencoder. The model unifies image generation and editing via flow matching: given a time \(t \in [0,1]\), a latent state \(z_t\), and conditioning \(c\), the network predicts the probability flow \(v_\phi(z_t, t, c)\) that transports a simple base distribution (e.g., Gaussian noise) to the data distribution. Training minimizes a squared-error objective between the predicted and ground-truth flow along a chosen interpolation path.

FLUX.1 Kontext supports in-context editing. The conditioning \(c\) can include both a text prompt and a reference image. We exploit this capability by providing the source-domain image \(x^{s}\) as the conditioning image and an instruction as the text conditioning to render the same scene in the target modality. In all experiments we use a fixed dataset-specific prompt, e.g., “Convert this to an IR image from the KAIST sensor” for KAIST and “Convert this to a SAR image” for M4-SAR, shared across all training pairs. We deliberately avoid image-specific captions; a detailed per-image textual description is an interesting direction for future work.

\subsection{LoRA-Adapted Flow Matching}

Directly fine-tuning FLUX.1 Kontext for each new spectral pair would require updating billions of parameters and substantial compute. Instead, we adopt Low-Rank Adaptation (LoRA), inserting small trainable matrices into selected layers while freezing the base model.

For a weight matrix \(W \in \mathbb{R}^{d_\text{out} \times d_\text{in}}\) in the attention or MLP projections of the transformer backbone, LoRA introduces a low-rank increment:
\[
W' = W + \Delta W, \quad \Delta W = \frac{\alpha}{r} A B,
\]
where \(A \in \mathbb{R}^{d_\text{out} \times r}\), \(B \in \mathbb{R}^{r \times d_\text{in}}\) are trainable, \(r \ll \min(d_\text{out}, d_\text{in})\) is the rank, and \(\alpha\) is a scaling factor. Only \(A\) and \(B\) are updated; all original weights in FLUX.1 Kontext remain fixed. We attach these LoRA adapters to the query, key, value, and output projections in each self-attention layer, as well as to the linear projections in the MLP sub-blocks of the image transformer backbone, yielding additional trainable parameters on the order of \(\sim 1\%\) of the base model while remaining expressive enough to capture the cross-spectral mapping in each domain.

In our experiments, we define a “training step” as a single optimizer update. We fine-tune LoRA using a batch size of 1. We adopt the standard LoRA initialization in which the initial effective weight update is \(\Delta W = 0\)~\cite{hu2022lora}. Specifically, we initialize the down-projection with Kaiming uniform and the up-projection to zeros~\cite{ostris_ai_toolkit_lora,peft_lora_docs}, ensuring training starts from the base model’s behavior. Investigating alternative LoRA initialization schemes, and their effects in low-data regimes, is a promising direction for future work.

\subsection{Model Selection via LPIPS}

We use LPIPS (Learned Perceptual Image Patch Similarity)~\cite{zhang2018lpips} as a model-selection metric. LPIPS measures the $\ell_2$ distance between deep feature representations of a synthetic image and its real counterpart, computed from a fixed AlexNet backbone, and correlates strongly with human perceptual judgments of image similarity. Lower LPIPS indicates closer perceptual match to the target modality. We specifically selected LPIPS over alternatives such as PSNR, SSIM, and Fréchet Inception Distance (FID) for two reasons. First, LPIPS has been shown to correlate more strongly with human perceptual similarity than PSNR or SSIM~\cite{zhang2018lpips}. This makes LPIPS more suitable for our cross-spectral translation setting, where preserving semantic structure and textures is more important than minimizing pixel-wise error. Second, we operate in an extremely low-data regime on the sensor-specific validation splits (50 paired images per domain), where distribution-level metrics such as FID become statistically fragile. 

For each dataset (KAIST and M4-SAR), we sweep a grid of LoRA hyperparameters. Let \(\lambda\) denote a LoRA hyperparameter configuration (learning rate, rank \(r\) with \(\alpha = r\), and number of training steps), and let \(\{\lambda_1, \dots, \lambda_n\}\) be the set of configurations in the sweep (with \(n = 15\) in our experiments):

\begin{itemize}
    \item learning rate \(\in \{1\times10^{-4}, 5\times10^{-4}\}\),
    \item rank $r \in \{16, 32\}$ with $\alpha = r$,
    \item training steps \(\in \{1\text{k}, 3\text{k}, 6\text{k}, 10\text{k}, 30\text{k}, 40\text{k}\}\).
\end{itemize}

For 1k, 3k, and 6k steps we train all \(2\times2\) combinations of learning rate and rank \(r\) (with \(\alpha = r\)), while for 10k, 30k, and 40k we instantiate only the configuration with learning rate \(5\times10^{-4}\) and rank \(r = 16\) (i.e., \(\alpha = 16\)), resulting in \(n = 15\) LoRA configurations \(\{\lambda_i\}_{i=1}^n\) per dataset, chosen to fit within our compute and time constraints.
Each configuration \(\lambda_i\) is trained on the \emph{Sensor Sample} split (100 paired images). We then:

\begin{enumerate}
    \item Translate the 50-image \emph{Sensor Val} split using each LoRA adapter.
    \item Compute LPIPS between each synthetic image and its real IR/SAR counterpart.
    \item Use the average LPIPS as a cross-spectral validation score.
\end{enumerate}

The configuration with lowest LPIPS is selected as the best LoRA for that dataset, which we denote by \(\lambda^\ast\). This avoids training detectors for every configuration and turns LoRA selection into a computationally efficient generative evaluation problem.

\subsection{Synthetic Dataset Construction for Detection}

Once the best LoRA is chosen:

\begin{itemize}
    \item For KAIST and M4-SAR, we translate every image in the \emph{Train} split from the source modality (RGB) into the target (IR/SAR), obtaining a synthetic training set \(\hat{\mathcal{D}}_\text{Train}\).
    \item Because translation is pixel-aligned, we reuse the original bounding boxes from the real \emph{Train} split without modification.
    \item For the cross-dataset KAIST experiments, we similarly translate RGB-only images from LLVIP and FLIR ADAS into the KAIST IR domain and reuse their labels.
\end{itemize}

Object detectors are then trained on real-only, synthetic-only, or real + synthetic combinations, depending on the experiment.

\section{Experimental Setup}

\subsection{Datasets and Splits}

\subsubsection{KAIST Multispectral Pedestrian}

KAIST contains 95k RGB--IR images captured from a vehicle-mounted beam splitter, with annotations for person, people, and cyclist classes~\cite{hwang2015kaist}. We focus on the \emph{person} class and restrict our experiments to daytime sequences to avoid dark RGB frames that may induce hallucinated structures during translation. We define five non-overlapping splits:

\begin{itemize}
    \item \textbf{Sensor Sample} (100 pairs): unlabeled, pixel-aligned RGB--IR pairs used exclusively for LoRA training.
    \item \textbf{Sensor Val} (50 pairs): unlabeled RGB--IR pairs used for LPIPS-based LoRA selection.
    \item \textbf{Train} (800 pairs): pixel-aligned RGB--IR pairs with bounding boxes used for detector training.
    \item \textbf{Val} (200 images): IR-only frames with annotations used for model selection.
    \item \textbf{Test} (911 images): IR-only frames with annotations used for final evaluation.
\end{itemize}
These splits represent a small, curated subset of the full KAIST corpus (which contains 95k RGB--IR images), chosen to emulate a data-scarce setting. KAIST is organized into numbered driving sequences; we use sequence~1 exclusively for \emph{Sensor Sample} and \emph{Sensor Val}, sequences~0 and~2 for \emph{Train} and \emph{Val}, and sequences~9--11 for \emph{Test}, ensuring no frame overlap across splits.

\subsubsection{External RGB Datasets: LLVIP and FLIR ADAS}
\label{subsec:kaist_cross}

For cross-dataset scaling, we leverage two additional multispectral pedestrian datasets but use only their RGB views during training:

\begin{itemize}
    \item \textbf{LLVIP}: 30{,}976 images (15{,}488 RGB--thermal pairs) from a binocular RGB--IR sensor in low-light conditions, with pedestrian bounding boxes~\cite{llvip}.
    \item \textbf{FLIR ADAS}: 26{,}000 RGB--thermal pairs at \(640\times512\) resolution, with more than 520k bounding boxes over 15 categories (person, bicycle, car, bus, etc.)~\cite{flir}.
\end{itemize}

We discard their IR channels, translate the RGB images into the KAIST IR domain using the best KAIST LoRA, and reuse the original labels. The resulting synthetic frames are appended to the KAIST \emph{Train} split to test whether external RGB corpora can be reused for IR detection via cross-spectral translation.

\begin{figure*}[!tb]
    \centering
    \includegraphics[
        width=0.9\textwidth,
        trim=0 20 0 0,
        clip
    ]{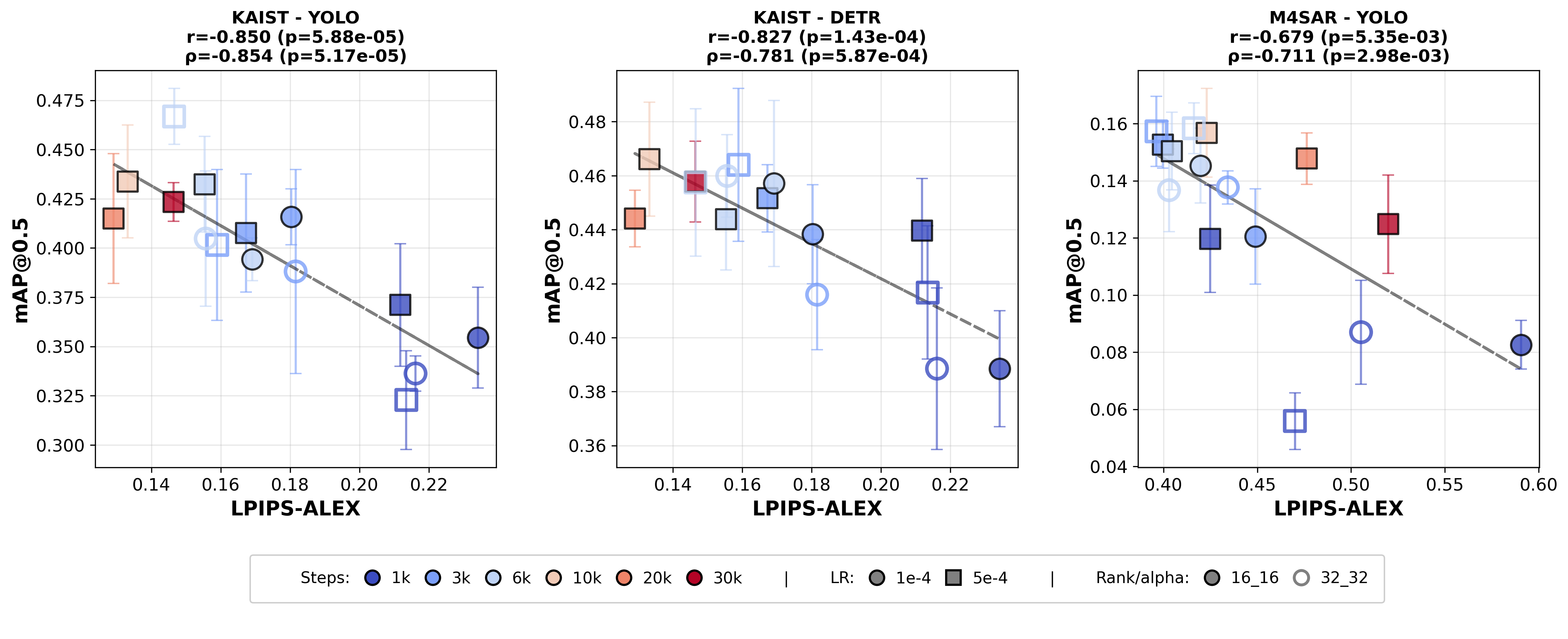}
    \caption{LPIPS on \emph{Sensor Val} versus YOLOv11n / DETR mAP@0.50 on the real \emph{Test} sets. From left to right: (i) KAIST with YOLOv11n, (ii) KAIST with DETR, and (iii) M4-SAR with YOLOv11n. Each point corresponds to a LoRA configuration: color encodes the number of LoRA training steps, marker shape encodes learning rate, and filled versus unfilled markers encode LoRA rank. Points show the mean over 5 runs per configuration and error bars indicate \(\pm 1\) standard deviation. Solid lines show least-squares linear fits. Panel titles report Pearson (r) and Spearman (\(\rho\)) correlation coefficients with associated p-values, all indicating strong negative correlations between LPIPS and downstream detection performance.}
    \label{fig:lpips_corr}
\end{figure*}

\subsubsection{M4-SAR RGB--SAR Benchmark}

M4-SAR provides co-registered RGB--SAR image pairs at 10\,m (VH) and 60\,m (VV) resolution. We use only the 10\,m VH subset (files 1--56087) and restrict detection to two classes: \emph{bridge} and \emph{harbor}.

From the official train and test partitions we construct:

\begin{itemize}
    \item \textbf{Sensor Sample} (100 pairs): unlabeled, pixel-aligned RGB--SAR pairs used exclusively for LoRA training.
    \item \textbf{Sensor Val} (50 pairs): unlabeled, pixel-aligned RGB--SAR pairs used for LPIPS-based LoRA selection.
    \item \textbf{Train} (1600 pairs): pixel-aligned RGB--SAR pairs with bounding boxes used for detector training.
    \item \textbf{Val} (400 images): SAR-only frames with annotations used for model selection.
    \item \textbf{Test} (200 images): SAR-only frames with annotations used for final evaluation.
\end{itemize}

The dataset is highly imbalanced: bridges account for roughly 94--96\% of all annotations across splits, with harbors making up the remainder. Similar to KAIST, our dataset partitions form a modest subset of the full M4-SAR dataset, reflecting a practical regime where high-resolution SAR annotations are scarce.

\subsection{LoRA Training and Synthetic Generation}

For each combination of hyperparameters in our grid, we fine-tune LoRA modules on the \emph{Sensor Sample} split of the corresponding dataset using the flow-matching loss inherited from FLUX.1 Kontext and condition on the source image and a fixed dataset-specific instruction prompt described in Section~3.2. After training:

\begin{enumerate}
    \item We translate the 50 \emph{Sensor Val} pairs and compute mean LPIPS against real IR/SAR images.
    \item We translate the \emph{Train} split (800 KAIST pairs / 1600 M4-SAR pairs) to obtain synthetic IR/SAR training sets.
\end{enumerate}

For all detection experiments, we train and evaluate detectors exclusively on target-modality images (IR or SAR) and their ground-truth bounding boxes. RGB images are used only as inputs to the translation model to generate synthetic target-modality data. All synthetic images are pixel-aligned with their corresponding source RGB frames.

\subsection{Object Detection Models}

We study two object detection families:

\begin{itemize}
    \item \textbf{YOLOv11n} (Ultralytics): a modern one-stage detector optimized for efficiency. We train for 30 epochs with batch size 16 and default Ultralytics hyperparameters.
    \item \textbf{DETR}: a transformer-based detector representing a conceptually different architecture. We train for 30 epochs with learning rate \(1\times10^{-5}\) and batch size 8.
\end{itemize}

Unless otherwise noted, detectors are trained on either real or synthetic versions of the \emph{Train} split and evaluated on real \emph{Test} images only. For each training configuration, we report mean and standard deviation of mAP over five independent train and test runs using different random seeds.

\section{Results}

\subsection{LPIPS as a Proxy for Downstream Detection}

For each dataset and LoRA hyperparameter configuration, we:

\begin{enumerate}
    \item Compute the average LPIPS on \emph{Sensor Val} between synthetic and real images.
    \item Train YOLOv11n (and DETR for KAIST) on the corresponding synthetic \emph{Train} set.
    \item Evaluate mAP on the real \emph{Test} split.
\end{enumerate}

Figure~\ref{fig:lpips_corr} visualizes LPIPS versus mAP@0.50 across all LoRA configurations for KAIST (YOLOv11n and DETR) and M4-SAR (YOLOv11n).

\begin{figure*}[!t]
    \centering
    \includegraphics[width=0.99\textwidth]{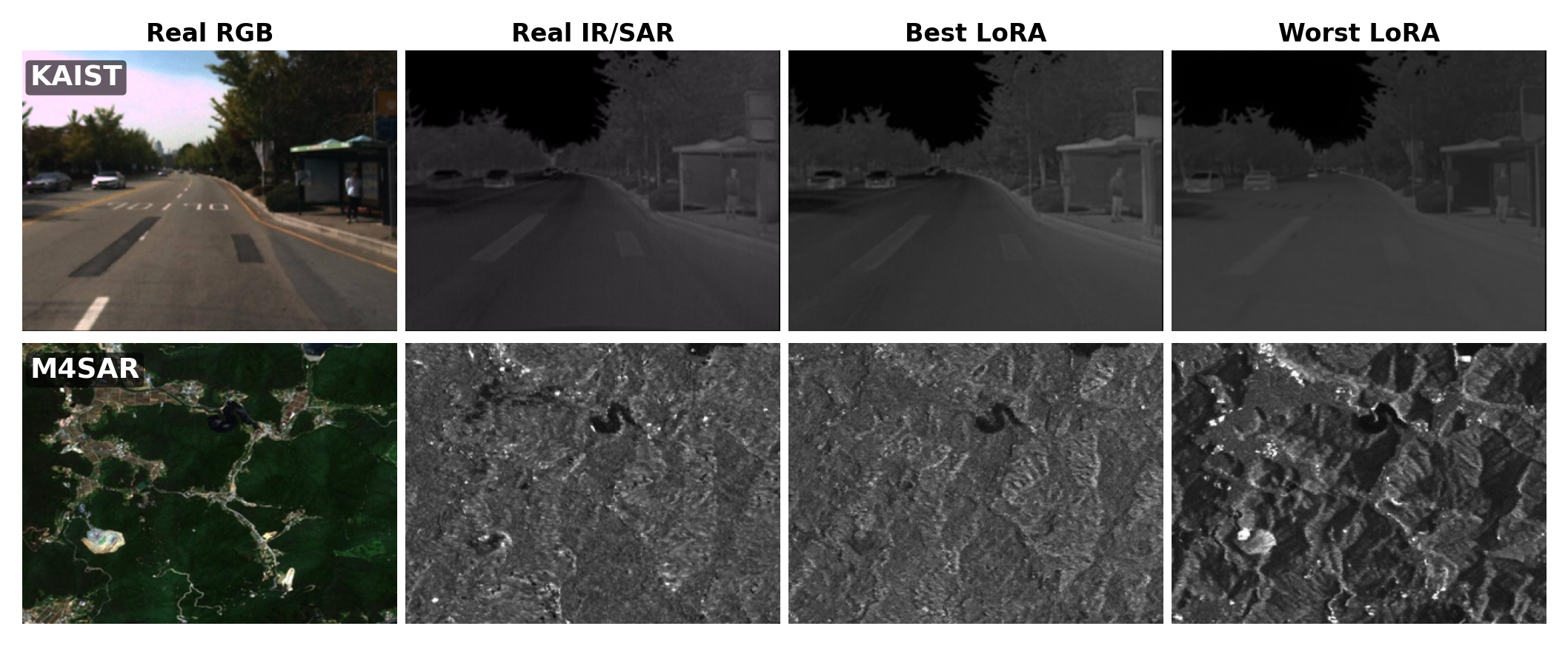}%
    \caption{Examples of cross-modal image translation on the KAIST (RGB--IR) and M4-SAR (RGB--SAR) datasets. For each dataset and scene, columns show the input RGB image, the corresponding real IR/SAR image, and synthetic IR/SAR images generated by the best- and worst-performing LoRA configurations, ordered from left to right. Best and worst are ranked by downstream YOLOv11n mAP@0.50.}
    \label{fig:qualitative_examples}
\end{figure*}

\begin{table*}[!t]
\centering
\caption{mAP on the KAIST \emph{Test} set for YOLOv11n and DETR with and without synthetic IR translated from external RGB datasets.}
\label{tab:crossdataset}
\setlength{\tabcolsep}{6pt}
\small
\begin{tabular}{lcccc}
\toprule
& \multicolumn{2}{c}{\textbf{YOLOv11n}} & \multicolumn{2}{c}{\textbf{DETR}} \\
\cmidrule(lr){2-5}
\textbf{Train Set} & \footnotesize{@0.50} & \footnotesize{@[0.5:0.95]} & \footnotesize{@0.50} & \footnotesize{@[0.5:0.95]} \\
\midrule
Real KAIST (baseline)   & $0.50 \pm 0.02$ & $0.22 \pm 0.01$ & $0.48 \pm 0.02$ & $0.19 \pm 0.01$ \\
+ FLIR-synth (400 imgs) & $\mathbf{0.54 \pm 0.02}$ & $0.23 \pm 0.01$ & $0.47 \pm 0.02$ & $0.18 \pm 0.01$ \\
+ LLVIP-synth (500 imgs)& $\mathbf{0.54 \pm 0.02}$ & $0.22 \pm 0.01$ & $0.48 \pm 0.01$ & $0.19 \pm 0.00$ \\
\bottomrule
\end{tabular}
\end{table*}

For KAIST and YOLOv11n (left panel of Fig.~\ref{fig:lpips_corr}), each point corresponds to a LoRA configuration. We observe a clear negative correlation: models with lower LPIPS on \emph{Sensor Val} achieve higher mAP@0.50 on KAIST \emph{Test}. The linear fit captures this trend quantitatively, with Pearson correlation \(r=-0.85\) (\(p=5.88\times10^{-5}\)) and Spearman correlation \(\rho=-0.85\) (\(p=5.17\times10^{-5}\)).

The middle panel of Fig.~\ref{fig:lpips_corr} shows the same behavior for DETR on KAIST, again with a strong negative relationship between LPIPS and downstream mAP (Pearson \(r=-0.83\), Spearman \(\rho=-0.78\), both highly significant).

In the right panel of Fig.~\ref{fig:lpips_corr}, the same pattern arises in the RGB-to-SAR setting on M4-SAR: lower LPIPS corresponds to higher detection mAP on the \emph{Test} set, despite class imbalance and domain complexity (Pearson \(r=-0.68\) (\(p=5.35\times10^{-3}\)), Spearman \(\rho=-0.71\) (\(p=2.98\times10^{-3}\))).

Across both datasets and architectures, low LPIPS on just 50 validation pairs reliably predicts which LoRA configuration yields the best downstream detection performance. Practically, this means we can select a LoRA adapter without training detectors for all configurations, drastically reducing search cost.

Figure~\ref{fig:qualitative_examples} illustrates how these quantitative trends manifest visually. For both KAIST and M4-SAR, the best LoRA (ranked by YOLOv11n mAP@0.50) produces IR/SAR images whose global contrast and local structures more closely match the real sensors, particularly around pedestrians, bridges, and harbor structures. The worst-performing LoRA, in contrast, exhibits blurrier backgrounds, distorted object shapes, and spurious textures, which likely reduce the utility of these images for detector training.

\subsection{Cross-Dataset Extension on KAIST}

We next ask whether the best KAIST LoRA can be used to translate external RGB corpora into KAIST-style IR and thereby extend the effective IR training set. Using the best KAIST adapter (chosen via LPIPS), we translate 400 FLIR ADAS RGB frames and 500 LLVIP RGB frames into synthetic KAIST-like IR images and append them to the real KAIST \emph{Train} set. Table~\ref{tab:crossdataset} reports results:

\begin{itemize}
    \item For YOLOv11n, mAP@0.50 improves from $0.50$ (KAIST-only baseline) to $0.54$ with FLIR-synth and $0.54$ with LLVIP-synth.
    \item For DETR, performance remains essentially unchanged or slightly degraded.
\end{itemize}

These gains are particularly notable because no additional IR annotations are required: we simply reuse labels from FLIR and LLVIP after translation. Qualitatively, synthetic IR images preserve pedestrian shapes and coarse scene layout while adapting contrast and background clutter to match KAIST’s thermal domain.

Taken together, these experiments show that a single LoRA-adapted flow-matching model can act as a practical translator that unlocks RGB-only datasets for IR detection. In terms of perceptual quality, the LPIPS of our best KAIST LoRA (0.129 on \emph{Sensor Val} using only 100 aligned RGB--IR training pairs in \emph{Sensor Sample}) is on par with the strongest Physics-Informed Diffusion (PID) configuration on KAIST (0.128 LPIPS), and sits near the bottom of the 0.37--0.13 LPIPS range reported for GAN and diffusion baselines in Table~1 of Mao et al.~\cite{mao2026pid}.

\subsection{Scaling SAR Detection with Synthetic SAR}
Finally, we investigate whether synthetic SAR generated from RGB images can boost detection performance on M4-SAR. Using the best M4-SAR LoRA (again chosen by \emph{Sensor Val} LPIPS), we translate 5000 RGB images into synthetic SAR and train YOLOv11n in three regimes:
\begin{enumerate}
    \item Real-only: 1600 real SAR images.
    \item Synthetic-only: 5000 synthetic SAR images.
    \item Real + Synthetic: 1600 real + 5000 synthetic.
\end{enumerate}

\begin{table}
\centering
\small
\setlength{\tabcolsep}{2.5pt}
\caption{M4-SAR: YOLOv11n detection performance with real and synthetic SAR.}
\label{tab:m4sar_scaling}
\begin{tabular}{lcccc}
\toprule
\textbf{Train Set} & \textbf{\# Real} & \textbf{\# Synth} &
\textbf{mAP@0.50} & \textbf{mAP@}\\[-2pt]
&&& & \textbf{[0.50:0.95]}\\
\midrule
Real-only            & 1600 & 0    & $0.19 \pm 0.01$ & $0.06 \pm 0.01$ \\
Synthetic-only       & 0    & 5000 & $0.18 \pm 0.02$ & $0.06 \pm 0.01$ \\
Real + Synthetic     & 1600 & 5000 & $\mathbf{0.25 \pm 0.01}$ & $\mathbf{0.09 \pm 0.01}$ \\
\bottomrule
\end{tabular}
\end{table}

Table~\ref{tab:m4sar_scaling} summarizes the results. Real-only provides the baseline, synthetic-only is slightly worse, and combining real and synthetic SAR yields a substantial boost in both mAP@0.50 and mAP@[0.50:0.95] (a \(>\)30\% relative gain over the real-only baseline at mAP@0.50).

Despite severe class imbalance and the complex statistics of SAR imagery, synthetic SAR clearly acts as an effective data augmenter when combined with limited real SAR.

\section{Discussion}

\subsection{LPIPS as a Practical Selection Signal}

Our experiments support a simple but powerful observation: LPIPS on a tiny validation set is a strong proxy for downstream object detection performance. LPIPS is computed on only 50 paired images per dataset, yet it predicts trends in mAP obtained from training full detectors on hundreds or thousands of translated images. This has practical implications:

\begin{itemize}
    \item LoRA selection can be guided by generative quality alone, avoiding expensive end-to-end detector retraining for each hyperparameter setting.
    \item In low-resource settings where detection labels are scarce (or available only for a subset of the domain), LPIPS offers an inexpensive surrogate for the task relevance of synthetic data.
\end{itemize}

Nevertheless, LPIPS is an image-level metric. It does not explicitly account for object-level fidelity or radiometric correctness, which may become important for fine-grained scientific applications.

\subsection{Synthetic Data as a Bridge Across Modalities}

Our results on KAIST and M4-SAR highlight complementary roles for synthetic data:

\begin{itemize}
    \item RGB$\rightarrow$IR translation enables cross-dataset expansion. One can harvest RGB-only datasets and map them into the IR domain, extending training distributions without new IR sensors or annotations.
    \item RGB$\rightarrow$SAR translation provides within-dataset scaling. Synthetic SAR increases sample diversity for underrepresented classes and viewpoints, boosting detector performance when combined with real SAR.
\end{itemize}

Importantly, synthetic-only training lags behind real-only baselines in the SAR case, underscoring that current generators exhibit a \textit{sim-to-real gap} and do not yet fully replace real measurements. Instead, they act as amplifiers for scarce real data.

\subsection{Architectural Sensitivity to Synthetic Data}

YOLOv11n consistently benefits from synthetic augmentation, especially on KAIST, whereas DETR shows minimal gains and occasionally slight regressions. Several factors may contribute:

\begin{itemize}
    \item One-stage detectors such as YOLOv11n may better exploit the improved background diversity and object variety introduced by synthetic data.
    \item DETR’s transformer architecture can be more data intensive and may require larger or more diverse datasets to realize benefits from augmentation.
    \item Synthetic artifacts (for example, subtle texture inconsistencies) might interact differently with each model’s inductive biases.
\end{itemize}

Exploring detector architectures tailored to synthetic-heavy regimes, or joint training of generator and detector, is an interesting direction for future work.

\subsection{Limitations and Future Directions}

We summarize limitations and directions for future work:

\begin{itemize}
    \item \textbf{Paired data requirement.} While we use only 100 co-registered pairs per modality, some settings may lack any paired data. A promising direction is to fine-tune a text-to-image backbone with LoRA using unpaired text supervision, then reuse the same adapters for image-to-image translation, reducing or eliminating the need for co-registered pairs.
    \item \textbf{Limited modalities and tasks.} We focus on RGB$\rightarrow$IR and RGB$\rightarrow$SAR for pedestrian and infrastructure detection. Other modalities such as radio-frequency (RF) imagery and tasks such as segmentation, tracking, and change detection remain to be explored. We also operate in a low-data regime (hundreds of training images), which may underestimate the gains achievable when combining foundation-model translation with large-scale labeled IR/SAR datasets.
    \item \textbf{Radiometric fidelity.} Our method optimizes for perceptual similarity rather than physical accuracy. For scientific remote sensing applications, enforcing sensor-specific radiometric constraints or leveraging physics-informed priors may be necessary.
    \item \textbf{Single foundation model.} All experiments use FLUX.1 Kontext as the base generator. Investigating other foundation models (e.g., text--vision or remote sensing FMs) and comparing adaptation strategies could reveal when flow-matching is most advantageous.
    \item \textbf{Detector scope.} We evaluate one lightweight one-stage detector (YOLOv11n) and one transformer-based detector (DETR). Extending this evaluation to additional architectures (including real-time and few-shot detectors) under the same data and compute constraints is an important direction for future work.
\end{itemize}

Despite these limitations, our results demonstrate a promising path forward: a single flow-matching foundation model, adapted via lightweight LoRA modules, can serve as a reusable cross-spectral translator that meaningfully improves IR and SAR detection in low-data regimes.

\newpage

{
    \small
    \bibliographystyle{ieeenat_fullname}
    \bibliography{main}
}

\end{document}